\documentclass[sigconf]{acmart}
\AtBeginDocument{%
  }

\copyrightyear{2026}
\acmYear{2026}
\setcopyright{cc}
\setcctype{by}
\acmConference[MM '26]{Proceedings of the 34th ACM International Conference on Multimedia}{November 10--14, 2026}{Rio de Janeiro, Brazil}
\acmBooktitle{Proceedings of the 34th ACM International Conference on Multimedia (MM '26), November 10--14, 2026, Rio de Janeiro, Brazil}
\acmDOI{10.1145/3767308.3835936}
\acmISBN{979-8-4007-2213-4/2026/11}

\acmSubmissionID{5722}

\usepackage{balance}
\usepackage{latexsym}
\usepackage[T1]{fontenc}
\usepackage[utf8]{inputenc}
\usepackage{microtype}
\usepackage{graphicx}
\usepackage{amsmath}
\usepackage{bm}
\usepackage{array}
\usepackage{makecell}
\usepackage{multirow}
\usepackage{float}
\usepackage{hyperref}
\usepackage{url}
\PassOptionsToPackage{prologue,dvipsnames, x11names, table}{xcolor}
\usepackage[dvipsnames]{xcolor}
\usepackage[x11names]{xcolor}
\usepackage[table]{xcolor}
\usepackage{xcolor}
\usepackage[most]{tcolorbox}
\usepackage{colortbl}
\usepackage{bbding}
\usepackage{pifont}
\usepackage{booktabs}
\usepackage{capt-of}
\usepackage{subfigure}
\usepackage{longtable}
\usepackage{titletoc}
\usepackage{arydshln} 
\usepackage{enumitem}
\usepackage{todonotes}
\usepackage{algorithm}
\usepackage{algpseudocode}
\usepackage{algcompatible}
\usepackage{CJKutf8}
\usepackage{algpseudocode}

\begin{document}

\title{Reinforcing Multimodal Reasoning via Token-Level Perception-Grounded Advantage Estimation}

\author{Zhihan Zhang}
\affiliation{%
  \institution{Singapore Management University}
  \city{Singapore}
  \country{Singapore}}
\email{zhihanzhang.2024@phdcs.smu.edu.sg}

\author{Lizi Liao}
\affiliation{%
  \institution{Singapore Management University}
  \city{Singapore}
  \country{Singapore}}
\email{lzliao@ smu.edu.sg}


\begin{abstract}
Reinforcement Learning with Verifiable Rewards (RLVR) has improved the reasoning capabilities of Multimodal Large Language Models (MLLMs), yet existing frameworks rely on coarse, sequence-level reward signals that lack the fine-grained supervision over the visually-grounded steps within a multimodal reasoning chain. 
We investigate this gap through the lens of two token-level metrics: \textit{visual dependency} (\textit{i.e.} how much a token's prediction relies on the input image features) and \textit{predictive entropy}.
Our empirical analysis reveals two key findings:
(1)~correct reasoning chains exhibit a markedly sharper entropy reduction as visual grounding intensifies, compared to incorrect ones;
(2)~pivotal tokens, those whose misprediction triggers reasoning collapse, are statistical outliers in the joint distribution of visual dependency and predictive entropy derived from correct chains.
Motivated by these findings, we propose \textbf{T}oken-level \textbf{P}erception-grounded \textbf{A}dvantage \textbf{E}stimation (\textbf{TPAE}), which estimates token-level advantages by measuring each token's statistical consistency with the vision-entropy patterns of correct rollouts. TPAE leverages this granular score to modulate the sequence-level advantage, producing a fine-grained supervision signal that can be integrated into various RLVR frameworks.
Extensive experiments on seven benchmarks show that TPAE consistently outperforms leading strong baselines, yielding more stable and efficient optimization for multimodal reasoning. The code is publicly available at \url{https://github.com/Zhihan72/TPAE}.

\end{abstract}




\maketitle

\section{Introduction}

Reinforcement Learning from Verifiable Rewards (RLVR), powered by efficient online algorithms such as Group Relative Policy Optimization (GRPO) \cite{shao2024grpo}, has fundamentally shifted the frontier of reasoning capabilities in Large Language Models (LLMs). 
By optimizing models using verifiable signals, such as structured thinking formats and final answer accuracy, RLVR has demonstrated strong empirical success in models like DeepSeek-R1 \cite{deepseekai2025deepseekr1incentivizingreasoningcapability}.
Translating this success to Multimodal Large Language Models (MLLMs), however, introduces unique complexities where fine-grained perception is tightly synthesized with multi-step logical deduction \cite{mathvista_lupan,mathvision_wangke,mathverse_zhangrenrui}.
This challenge has motivated an array of research aimed at adapting RLVR to multimodal reasoning, primarily through
data-centric enhancements \cite{deng2025openvlthinker,Yang_2025_ICCV_R1_Onevision,visionr1_huang,deepseekai2025deepseekr1incentivizingreasoningcapability}, 
reward modelling \cite{fan2025grit,liu2026visionreasoner,xia2025visionaryr1mitigatingshortcutsvisual,fan2025sophiavlr1reinforcingmllmsreasoning}, and 
rollout-level algorithmic adjustments \cite{liu2025noisyrollout,yao2025rsharevl,wang2025vlrethinker}.

\begin{figure}[t]
    \centering
    \includegraphics[width=\linewidth]{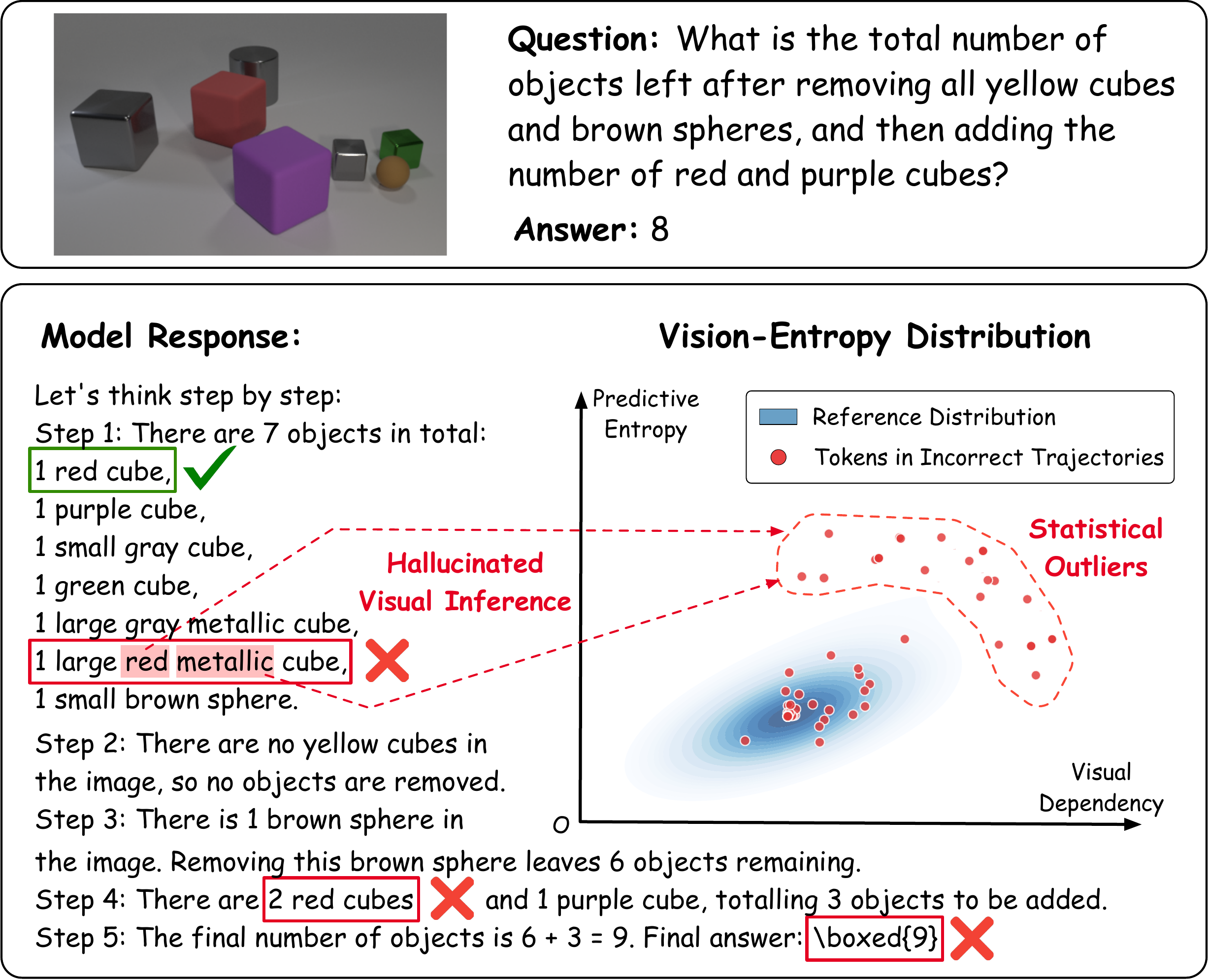}
    \caption{Token-level joint distribution of visual dependency and predictive entropy, where outlier tokens from incorrect trajectories emerge as the triggers of reasoning collapse.}
    \label{fig:task_case}
\end{figure}

Despite this progress, existing RLVR frameworks for multimodal reasoning are fundamentally limited by the coarse granularity of their reward signals \cite{wang2026perceptionaware,huang2026spotlight,xiao2025perceptionr1advancingmultimodalreasoning,zhu2026shuffler1efficientrlframework}.
Mainstream reinforcement learning (RL) algorithms such as GRPO \cite{shao2024grpo} and DAPO \cite{yu2025dapo} derive a single, rollout-level advantage from outcome-based rewards, assigning a uniform advantage value to every token within a rollout.
This undiscriminating assignment treats all tokens—whether reliably perception-grounded or responsible for reasoning failures—as equally contributing to the final answer, introducing gradient noise that hinders policy convergence \cite{lin2025critical,sun2025ktae}.
A natural remedy is to introduce finer-grained supervision via process-level reward models \cite{fan2025sophiavlr1reinforcingmllmsreasoning,xiao2025perceptionr1advancingmultimodalreasoning}; however, this approach requires training a separate learned scorer, which is data-demanding, computationally expensive, and vulnerable to reward hacking.
This motivates a key research question: \textit{Can we obtain fine-grained, perception-grounded advantage estimates directly from the policy's own reasoning behavior, without the overhead of auxiliary reward models?}

\begin{figure*}[t]
    \centering
    \includegraphics[width=0.98\linewidth]{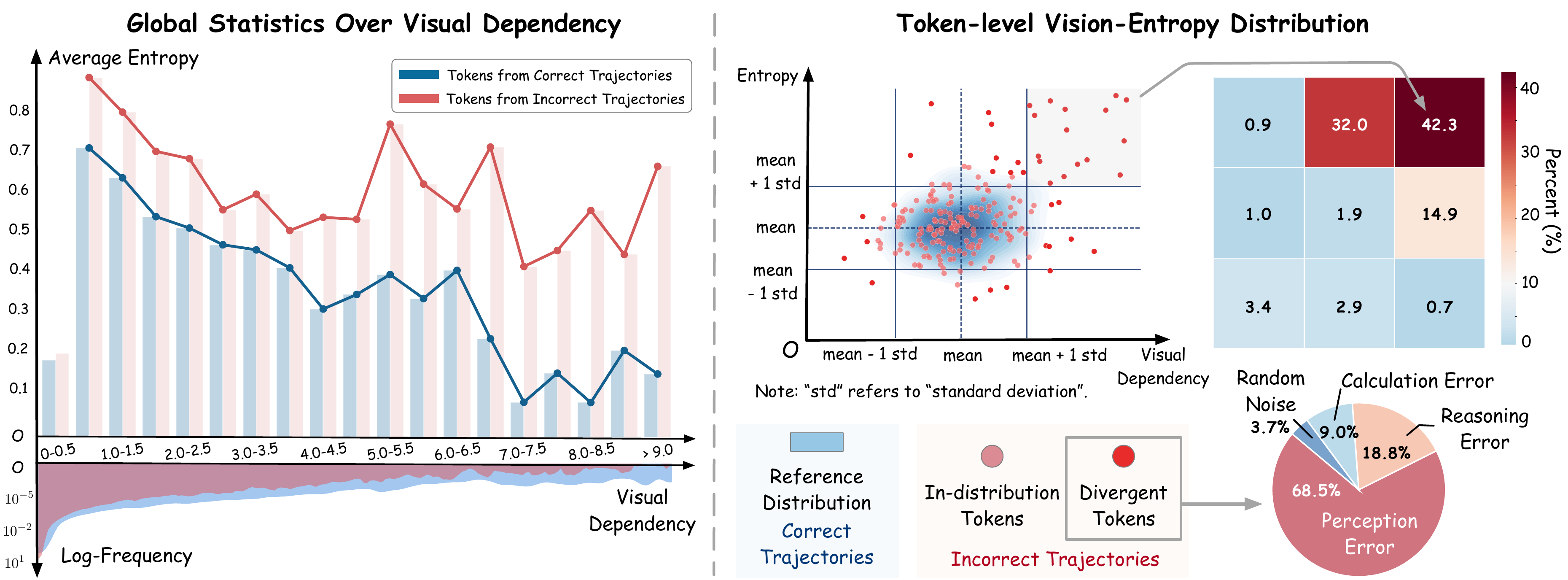}
    \caption{Empirical analysis of reasoning trajectories using Qwen2.5-VL-7B. (a) Distribution of average entropy and log-frequency across visual dependency levels (left). (b) Density of token-level vision-entropy misalignment relative to the reference distribution center (upper right). (c) Error analysis of divergent tokens through human annotation (lower right).}
    \label{fig:preliminary_charts}
\end{figure*}

To answer this question, we begin by examining two intrinsic metrics that are readily available from the policy model itself: a token’s visual dependency and its predictive entropy. Through an empirical analysis of 12,800 reasoning trajectories generated by Qwen2.5-VL-7B across 100 questions, we uncover a striking pattern illustrated in Figure~\ref{fig:preliminary_charts}: \textit{correct reasoning chains exhibit sharply lower entropy as visual dependency increases, whereas incorrect chains maintain high entropy despite strong visual dependency}—a phenomenon we term \textit{non-resolving grounding}, where the model attends to the visual input but fails to extract the information needed to disambiguate its next-token predictions. 
This widening gap between correct and incorrect chains establishes the coupling of visual dependency and predictive entropy as an effective, model-intrinsic indicator of multimodal reasoning quality.

To pinpoint the mechanistic drivers of this divergence, we zoom into the token-level distribution.
Using the joint distribution of visual dependency and entropy from correct reasoning chains as a reference, we measure the statistical alignment of each token in failed trajectories against the reference. While the majority of tokens in incorrect reasoning chains fall within the reference distribution, a critical 7.64\% emerge as severe outliers (\textit{e.g.}, the red-highlighted tokens in Figure~\ref{fig:task_case}). 
A rigorous human evaluation of 200 randomly sampled outliers confirms that \textit{these divergent tokens are not random noise but the semantic pivots that trigger visual mis-grounding and logical reasoning errors}—precisely the tokens that a fine-grained advantage estimator should down-weight.

Building on these findings, we propose \textbf{T}oken-level \textbf{P}erception-grounded \textbf{A}dvantage \textbf{E}stimation (\textbf{TPAE}), a novel algorithm that refines coarse, rollout-level reward signals into dense, token-level advantage estimates within multimodal RLVR frameworks—without training any auxiliary reward model.
TPAE jointly tracks each token's visual dependency and predictive entropy—its \textit{vision-entropy state}—and leverages correct rollouts to build reference distributions over these states.
Each token is scored by its statistical alignment with the corresponding reference distribution through a hypothesis testing perspective: tokens consistent with the reference are deemed trustworthy, while severe outliers are penalized.
These per-token scores then modulate the rollout-level advantage, directing gradient updates toward perceptually grounded reasoning and away from the triggers of reasoning failure.
As a lightweight, model-intrinsic module, TPAE can be seamlessly integrated into existing RL algorithms such as GRPO and DAPO. 

We validate the effectiveness of TPAE in advancing multimodal reasoning by integrating it into two base RL algorithms across two MLLM backbones. Our evaluation results across seven diverse benchmarks demonstrate that integrating TPAE with either GRPO or DAPO yields consistent performance gains; notably, our TPAE-D-Qwen2.5-7B surpasses existing state-of-the-art multimodal reasoning models utilizing the same Qwen2.5-VL-7B backbone.
To sum up, our \textbf{main contributions} are threefold:
\begin{itemize}[leftmargin=*,topsep=1pt]
\item We reveal that the coupling between token-level visual dependency and predictive entropy is an effective, model-intrinsic indicator of multimodal reasoning quality, and identify statistically divergent tokens as the drivers of reasoning failures.
\item We propose TPAE, a token-level perception-grounded advantage estimation algorithm that translates this insight into dense, per-token supervision requiring no auxiliary reward model.
\item Extensive experiments across seven multimodal benchmarks demonstrate the superior performance of TPAE over baseline algorithms and existing state-of-the-art models.
\end{itemize}

\section{Related Work}

\noindent \textbf{Multimodal Large Language Model.} 
Existing advancements in MLLMs have primarily focused on the modality alignment via multimodal projectors that bridge vision encoders with pre-trained Large Language Model (LLM) backbones. Diverse alignment strategies have been explored to optimize visual perception, ranging from compressing visual tokens \citep{Li2022BLIPBL, Bai2023QwenVLAF} to fusing multiple vision encoders \citep{NEURIPS2024_9ee3a664, lin2023sphinxjointmixingweights}. 
Beyond the exploration of architecture of MLLMs, large-scale instruction tuning has emerged as a central technique for improving multimodal alignment and overall performance on multimodal benchmarks \citep{liu2023visual, Bai2023QwenVLAF}. 
However, MLLMs still exhibit a significant performance gap in complex multimodal reasoning tasks \citep{qwen2_5-VL,qwen3technicalreport, mathvista_lupan,mathverse_zhangrenrui, zhang-etal-2025-xfinbench}. 
Drawing on the success of policy optimization in textual reasoning, recent works have begun adapting foundational algorithms such as PPO \cite{Schulman2017ProximalPO} and GRPO \cite{shao2024grpo} to the multimodal setting to tackle increasingly complex reasoning tasks.

\noindent \textbf{Multimodal Reinforcement Learning.}
Most strategies in multimodal reinforcement learning have focused on enhancing the RLVR framework from three perspectives: data, rollout and reward. 
Data-centric methods focus on the curation of large-scale, visually-grounded datasets by distilling chain-of-thought trajectories from proprietary models, typically employing a two-stage training paradigm consisting of initial supervised fine-tuning (SFT) followed by reinforcement learning via the GRPO objective \cite{deng2025openvlthinker,Yang_2025_ICCV_R1_Onevision,visionr1_huang,deepseekai2025deepseekr1incentivizingreasoningcapability, chen2025sft_VLAA_Thinking,huang2026visionr1incentivizingreasoningcapability_Vision_R1}.
Works on rollout improvements focus on diversifying the rollout space by visual augmentations \cite{liu2025noisyrollout,yao2025rsharevl} or sample replay mechanisms \cite{wang2025vlrethinker,leng2025mmr1enhancingmultimodalreasoning}.
Reward-centric works aim to evaluate the visual grounding of rollouts by either enforcing verifiable intermediate steps—such as the generation of spatial coordinates for relevant objects \cite{fan2025grit,liu2026visionreasoner} or auxiliary captioning tasks \cite{xia2025visionaryr1mitigatingshortcutsvisual}—or by deploying external neural reward models for dense supervision \cite{fan2025sophiavlr1reinforcingmllmsreasoning,Zhang_2025_ICCV_R1_VL}.

However, existing works predominantly conceptualize visual grounding as a post-hoc supervisory constraint, treating sampled rollouts as static artifacts for external verification. 
While more recent efforts have introduced token-level visual dependency to probe the internal dynamics of multimodal reasoning, this metric remains relegated to passive filters or heuristic penalties within the optimization objective \cite{wang2026perceptionaware, huang2026spotlight}.
In this work, we propose TPAE, a novel perception-grounded advantage estimation algorithm that leverages the token-level joint dynamics of visual dependency and predictive entropy to modulate the coarse reward signal.

\section{Preliminary}

Existing RLVR frameworks rely on coarse, sequence-level reward signals that overlook the varying significance of individual tokens. To move beyond this simplification, we introduce token-level entropy and visual dependency as granular metrics to characterize the internal dynamics of multimodal reasoning generation.
Our empirical analysis reveals that 
correct reasoning chains exhibit a sharper entropy reduction tied to visual grounding than the incorrect trajectories.
Furthermore, a non-trivial proportion of tokens from incorrect trajectories emerge as statistical outliers relative to the correct reference distribution. 
Through human evaluation, these outliers are confirmed to be the pivotal triggers responsible for multimodal reasoning collapse.

\subsection{Group Relative Policy Optimization}

Group Relative Policy Optimization (GRPO) \citep{shao2024grpo} is a variant of the Proximal Policy Optimization (PPO) \cite{Schulman2017ProximalPO} algorithm that removes the requirement for a separate value model by estimating advantages via group-based computation. 
For a given multimodal input $(I, q)$ consisting of a visual context $I$ and a textual query $q$, the old policy model $\pi_{\theta_{\text{old}}}$ samples a group of $G$ rollouts, $\{o_i\}_{i=1}^G$. 

In the RLVR framework, each rollout $o_i$ is assigned a binary reward $R_i \in \{0,1\}$, determined solely on whether its final answer matches the ground truth. The token-level advantage $\hat{A}_{i,j}$ is then defined as the sequence-level reward normalized across the group:
\begin{equation}
\hat{A}_{i,j} =\hat{A}_i = \frac{R_i - \text{mean}(\{R_k\}_{k=1}^G)}{\text{std}(\{R_k\}_{k=1}^G)}
\label{eq:adv_grpo}
\end{equation}
The policy $\pi_{\theta}$ is then updated to maximize a clipped surrogate objective, where this uniform advantage $\hat{A}_i$ is broadcast to every token $o_{ij}$ within the rollout $o_i$:
\begin{equation}
\begin{split}
    & \mathcal{L}^{\text{GRPO}}(\theta)  = \mathbb{E}_{[\{o_i\}_{i=1}^G\sim \pi_{\theta_{\text{old}}}(\cdot \mid I,q) ]}  \frac{1}{G} \sum_{i=1}^{G} \frac{1}{|o_i|} \sum_{j=1}^{|o_i|} \Bigl\{ \\
    & \min \left( r_{i,j}(\theta) \hat{A}_{i,j}, \text{clip}(r_{i,j}(\theta), 1 - \varepsilon, 1 + \varepsilon) \hat{A}_{i,j} \right) - \beta \mathbb{D}_{\text{KL}[\pi_\theta \mid \pi_{\text{ref}}]} \Bigl\}
\end{split}
\end{equation}
where $r_{i,t}(\theta) = \frac{\pi_{\theta}(o_{i,j} | I, q, o_{i,<j})}{\pi_{\theta_{\text{old}}}(o_{i,j} | I, q, o_{i,<j})}$ is the probability ratio.

Decoupled Clip and Dynamic Sampling Policy Optimization (DAPO) \citep{yu2025dapo} is a notable extension of GRPO, incorporating several modifications such as clip-higher, dynamic sampling, and token-level policy gradient loss (see supplementary materials for details). In this work, we investigate the application of our method to both GRPO and DAPO algorithms.

\begin{figure*}[t]
    \centering
    \includegraphics[width=\linewidth]{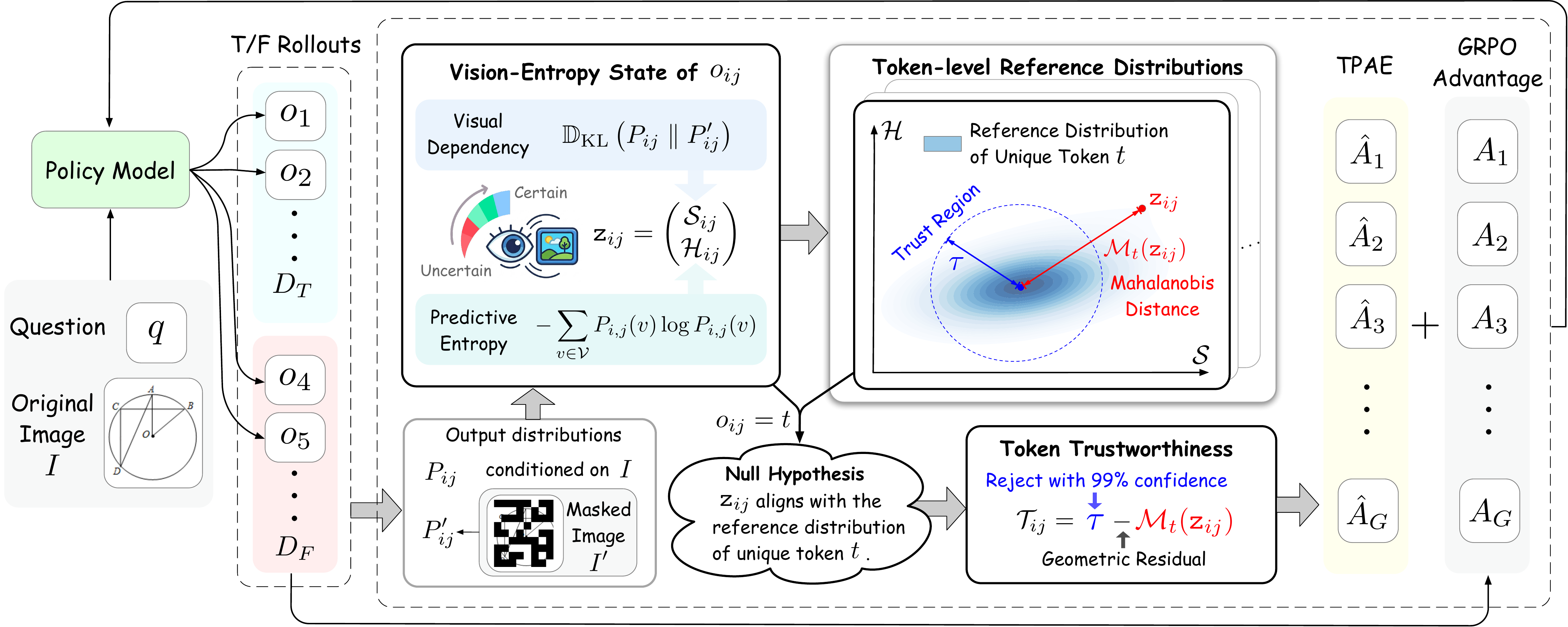}
    \caption{Overview of the TPAE algorithm. 
    TPAE utilizes correct rollouts to establish reference distributions of vision-entropy states of unique tokens. 
    From a hypothesis testing perspective, TPAE then applies a trust boundary $\tau$ to quantify token-level trustworthiness based on distributional alignment. 
    This metric is then integrated with the rollout-level advantage from GRPO.}
    \label{fig:algorithm}
\end{figure*}

\subsection{Token Entropy and Visual Dependency}

While GRPO mitigates reward sparsity through a group-relative advantage estimation, it relies on a coarse, sequence-level signal that ignores the varying significance of individual tokens. 
To move beyond this sequence-level simplification, we introduce token entropy and visual dependency as granular metrics to characterize the internal dynamics of the multimodal generation process.

\subsubsection{Predictive Entropy} 
The token-level entropy quantifies the model's predictive uncertainty in the reasoning chain. 
Specifically, for token $o_{ij}$ in rollout $o_i$, the policy model generates a probability distribution $P_{ij}$ over the vocabulary $\mathcal{V}$, i.e. $P_{i,j}=\pi_\theta(\cdot \mid I,q,o_{i,<j})$. The entropy of this prediction is then calculated as
\begin{equation}
\mathcal{H}_{i,j} := -\sum_{v \in \mathcal{V}} P_{i,j}(v) \log P_{i,j}(v)
\end{equation}
where $P_{i,j}(v)$ represents the likelihood of choosing token $v$.
A low $\mathcal{H}_{i,j}$ value indicates that the model is highly confident, with its probability concentrated on a single optimal next token. In contrast, a high $\mathcal{H}_{i,j}$ value indicates a state of high uncertainty, where the model identifies multiple plausible paths for the next step.

\subsubsection{Visual Dependency}
The token-level visual dependency measures the degree of reliance on the visual input $I$ during the generation of token $o_{ij}$ \cite{wang2026perceptionaware, huang2026spotlight}. 
Formally, we define $P'_{i,j}=\pi_\theta(\cdot \mid I',q,o_{i,<j})$ as the probability distribution over the vocabulary $\mathcal{V}$ conditioned on a masked, non-informative version of the image $I'$. 
From the perspective of information gain, the visual dependency is calculated as the Kullback-Leibler (KL) divergence between the original distribution $P_{i,j}$ and the vision-blind distribution $P'_{i,j}$:
\begin{equation}
\mathcal{S}_{ij} := \mathbb{D}_{\text{KL}} \left( P_{i,j} \parallel P'_{i,j} \right)= \sum_{v \in \mathcal{V}} P_{i,j}(v) \log \frac{P_{i,j}(v)}{P'_{i,j}(v)}
\end{equation}
A high $\mathcal{S}_{ij}$ value indicates that the visual input provides critical information for the prediction of token $o_{ij}$, marking it as a key moment of perception-anchored reasoning. 
Conversely, a low $\mathcal{S}_{ij}$ value suggests the model is relying primarily on linguistic knowledge or the preceding textual context to determine the next token.

Following \cite{wang2026perceptionaware,huang2026spotlight}, we implement a patch-based masking strategy for $I'$ at a 60\% masking ratio. Specifically, 60\% of the image patches are randomly sampled and masked, effectively restricting the available visual cues for visual dependency quantification.

\subsection{Analysis of Multimodal Reasoning}

We conduct an empirical investigation of multimodal reasoning using Qwen2.5-VL-7B-Instruct \cite{qwen2_5-VL} across 100 questions sampled from the ViRL39K dataset \cite{wang2025vlrethinker}. By generating 128 independent trajectories per question, we establish a robust distribution consisting of 5,995 correct and 6,805 incorrect trajectories.
We observe a clear structural divergence: correct reasoning chains are significantly more concise, averaging 418.07 tokens compared to the 546.57 tokens of their incorrect counterparts. 
Moreover, as shown in Figure~\ref{fig:preliminary_charts}(a), correct reasoning chains show a higher log-frequency density in regions of high visual dependency, while incorrect trajectories are heavily concentrated in the minimal dependency interval.

By integrating predictive entropy $\mathcal{H}_{ij}$ with visual dependency $\mathcal{S}_{ij}$, we investigate how visual grounding modulates the model's internal certainty across both correct and incorrect reasoning chains:

\textbf{Insight 1: Correct reasoning chains exhibit a sharper entropy reduction through visual grounding.}
Figure~\ref{fig:preliminary_charts}(a) illustrates the average entropy at varying levels of visual dependency, revealing a clear divergence in how visual information impacts model certainty. In correct trajectories, we observe a monotonic decrease in entropy as visual dependency increases, suggesting that the model successfully translates visual cues into predictive certainty. Conversely, in incorrect trajectories, entropy remains at a consistently high plateau regardless of the level of visual reliance. This divergence implies that while correct reasoning chains use visual context to effectively prune the output distribution, incorrect reasoning paths suffer from non-resolving grounding—a phenomenon where the model attends to the image but fails to extract the information required to resolve its internal uncertainty. 

\textbf{Insight 2: Vision-entropy misalignment pinpoints the pivotal tokens triggering multimodal reasoning collapse.}
We then characterize the token-level vision-entropy state $(\mathcal{S}_t, \mathcal{H}_t)$ by constructing the empirical joint distribution of visual dependency and entropy for each unique token $t$ across all trajectories of a given question.
By establishing the distributions from correct reasoning chains as a reference baseline, we can evaluate the statistical alignment of tokens generated in failed trajectories.
Our analysis reveals that while most tokens in failed trajectories align with the reference, a critical 7.64\% fall into the extreme distributional tails, deviating by more than three standard deviations from reference center.
To investigate these divergent tokens, we partition them into a $3 \times 3$ diagnostic grid, categorized by whether their predictive entropy and visual dependency deviate by more than one standard deviation from the reference mean. As illustrated in Figure~\ref{fig:preliminary_charts}(b), over 74\% of these tokens are concentrated within the upper-middle and upper-right regions of the grid. 
This concentration reveals a decoupling of perception and reasoning in failed trajectories, providing granular validation for the global statistics observed in Figure~\ref{fig:preliminary_charts}(a).

To validate these statistical trends, we conduct a human evaluation on a random subset of 200 divergent tokens, each drawn from a unique incorrect trajectory.
We tasked three annotators with classifying these tokens into four distinct categories.
As shown in Figure~\ref{fig:preliminary_charts}(c), 68.5\% of these tokens are perception errors stemming from visual mis-grounding, while 18.8\% are reasoning errors that misinterpret valid visual evidence during multi-step inference. Only 3.7\% represent benign noise, confirming that these divergent tokens are overwhelmingly linked to reasoning failure.
Further details on annotation protocol are provided in supplementary materials.
Collectively, these findings demonstrate that distributional divergence from reference distributions is an effective indicator for localizing the pivotal moments of multimodal reasoning collapse.

\section{Method}

Building upon our empirical findings, we propose \textbf{T}oken-level \textbf{P}erception-grounded  \textbf{A}dvantage \textbf{E}stimation (\textbf{TPAE}), an algorithm designed to refine coarse, sequence-level signals of outcome-based rewards into dense, token-level supervision by coupling visual dependency with predictive entropy.
For each question, TPAE leverages correct rollouts to establish reference distributions of vision-entropy states for unique tokens. 
TPAE then quantifies the trustworthiness of individual tokens by measuring their statistical alignment with these reference distributions through a hypothesis-testing perspective.
This metric is integrated directly into the rollout-level advantage computed by mainstream RL algorithms to penalize tokens deviating from the established trust region.

\subsection{Vision-Entropy Distribution}

For a given multimodal input $(I,q)$, we sample a group of $G$ rollouts, $\{o_i\}_{i=1}^G$, each with a corresponding rule-based, outcome-oriented reward $\{R_i\}_{i=1}^G$ indicating its correctness. We divide these rollouts into a correct set $D_T=\{o_i \mid R_i=1\}$ and an incorrect set $D_F=\{o_i \mid R_i=0\}$.
In our work, we utilize the vision-entropy distribution from correct rollouts $D_T$ as a reference baseline to capture the optimal joint dynamics of visual grounding and predictive entropy that facilitate a correct final answer.

For each unique token $t$ in the correct set $D_T$, we aggregate its observed vision-entropy states $Z_t=\{\mathtt{z}_{ij} \mid o_{ij}=t, o_i \in D_T\}$, where each state $\mathtt{z}_{ij}=[\mathcal{S}_{ij}, \mathcal{H}_{ij}]^\top \in \mathbb{R}^2$ captures the token's visual dependency $\mathcal{S}$ and entropy $\mathcal{H}$.
We then approximate the reference distribution for each unique token $t$ through multivariate Gaussian, i.e. $P(\mathtt{z}\mid t) \sim \mathcal{N}(\mathbf{\mu}_t, \mathbf{\Sigma}_t)$, defined as:
\begin{equation}
P(\mathtt{z} | \mathcal{\mu}_t, \mathbf{\Sigma}_t) = \frac{1}{2\pi\sqrt{ | \mathbf{\Sigma}_t|}} \exp\left( -\frac{1}{2} (\mathtt{z} - \mathcal{\mu}_t)^\top \mathbf{\Sigma}_t^{-1} (\mathtt{z} - \mathcal{\mu}_t) \right)
\label{eq:gaussian}
\end{equation}
where $\mathcal{\mu}_t = \frac{1}{|Z_t|}\sum_{\mathtt{z} \in Z_t}\mathtt{z}$, $\mathbf{\Sigma}_t=\frac{1}{|Z_t|-1}\sum_{\mathtt{z} \in Z_t} (\mathtt{z} - \mathcal{\mu}_t)^T(\mathtt{z} - \mathcal{\mu}_t)$.
The choice of a multivariate Gaussian is motivated by the observed unimodal centrality of vision-entropy states  in $D_T$ (see supplementary materials for comparisons with alternative parametric models).

To ensure the robustness of the token-level estimates, we compute the covariance matrix $\mathbf{\Sigma}_t$ using the Ledoit-Wolf adaptive shrinkage approach \cite{Ledoit2004AWE}.
This addresses the sample scarcity inherent to RLVR frameworks, where limited rollouts may yield unstable or singular token-level empirical estimates.
Specifically, we parameterize the estimated covariance $\hat{\mathbf{\Sigma}}_t$ as a convex combination of the empirical covariance $\mathbf{\Sigma}_t$ and a diagonal prior $\mathbf{\Sigma}_{t}^{\text{prior}}$, as follows:
\begin{equation}
\hat{\mathbf{\Sigma}}_t = (1-\lambda_t)\mathbf{\Sigma}_t+\lambda_t\mathbf{\Sigma}_{t}^{\text{prior}}
\label{eq:shrinkage}
\end{equation}
where $\mathbf{\Sigma}_{t}^{\text{prior}}=\text{diag}(\sigma^2_{\mathcal{S}_t},\sigma^2_{\mathcal{H}_t})$ denotes the marginal variances of visual dependency and entropy for token $t$, as estimated from the correct rollout set $D_T$. 
The shrinkage intensity $\lambda_t \in [0,1]$ follows the optimal shrinkage principle, $\lambda_t=\frac{3}{|Z_t|+3}$, which adaptively weights the prior based on the sample size $|Z_t|$ \cite{Ledoit2004AWE}. 
Further theoretical justifications are provided in the supplementary material.

\subsection{Token Trustworthiness Quantification}

We quantify the trustworthiness of individual tokens by measuring their statistical alignment with the reference distribution of correct reasoning paths.
Formally, we frame this quantification as a hypothesis test: for a token $o_{ij}=t$, the null hypothesis $H_0$ posits that its vision-entropy state $\mathtt{z}_{ij}=[\mathcal{S}_{ij}, \mathcal{H}_{ij}]^\top$ originates from the reference distribution $P(\mathtt{z} | \mathcal{\mu}_t, \mathbf{\Sigma}_t)$ derived from correct rollouts. 

To test the null hypothesis $H_0$, we calculate the likelihood of observing the specific vision-entropy state $\mathtt{z}_{ij}$ under the reference distribution $P(\mathtt{z} \mid \mu_t, \Sigma_t)$. For a multivariate Gaussian model, this probability density is strictly proportional to the exponentiated negative squared Mahalanobis distance \cite{mahalanobis_dist}:
\begin{equation}
P(\mathtt{z}_{ij} \mid \mu_t, \Sigma_t) \propto \exp \left( -\frac{1}{2} \mathcal{M}_{t}^2(\mathtt{z}_{ij}) \right)
\end{equation}
where $\mathcal{M}_{t}(\mathtt{z}_{ij}) = \sqrt{(\mathtt{z}_{ij} - \mu_t)^\top \mathbf{\Sigma}_t^{-1} (\mathtt{z}_{ij} - \mu_t)}$ denotes the Mahalanobis distance.
A larger $\mathcal{M}_{t}(\mathtt{z}_{ij})$ value corresponds to a lower likelihood that $\mathtt{z}_{ij}$ aligns with the reference distribution, providing a clear statistical signal to reject the null hypothesis $H_0$.

We employ a principled boundary $\tau$ to demarcate the trust region for $H_0$, leveraging the statistical properties of the Mahalanobis distance. 
Under $H_0$ where $\mathtt{z}_{ij} \sim \mathcal{N}(\mu_t, \Sigma_t)$, the squared Mahalanobis distance strictly follows a Chi-squared distribution with two degrees of freedom, i.e. $\mathcal{M}^2_{t}(\mathtt{z}_{ij}) \sim \chi^2_2$. 
By adopting a 99\% confidence level, we reject $H_0$ if the squared Mahalanobis distance exceeds the value $\chi^2_{2,0.99} \approx 9.21$. This threshold implies that a token's vision-entropy state $\mathtt{z}_{ij}$ is considered misaligned with the reference distribution if its deviation from the distribution center exceeds $\sqrt{9.21} \approx 3.03$ standard deviations, i.e., $\mathcal{M}_{t}(\mathtt{z}_{ij}) > \tau = 3.03$.

Building upon the above derivations, we formally quantify the trustworthiness $\mathcal{T}_{ij}$ of a token’s vision-entropy state as the residual distance to the established trust boundary:
\begin{equation}
\mathcal{T}_{ij} = \tau - \mathcal{M}_t(\mathtt{z}_{ij})
\label{eq:trust}
\end{equation}
Within this formulation, a positive trustworthiness score ($\mathcal{T}_{ij} \geq 0$) confirms that the token resides safely within the 99\% confidence interval of the reference distribution. Conversely, a negative score ($\mathcal{T}_{ij} < 0$) identifies a statistically significant outlier, signaling a rejection of the null hypothesis $H_0$ and a critical vision-entropy misalignment of token $o_{ij}$.
For tokens that are absent from the reference set $D_T$, we assign a default trustworthiness score of zero to maintain a conservative estimation.

\begin{table*}[t]
    \centering
    \caption{Main results (avg \@8 acc \%) across seven multimodal reasoning benchmarks. All evaluations utilize exact-match scoring on verifiable instances to ensure objective results, avoiding any LLM-as-a-judge. All comparative baselines are instantiated from the Qwen2.5-VL-7B backbone. Best performance of 7B models is marked with bold, second best with underline. }
    \vspace{-0.1cm}
    \label{table:experimental_result}
    \setlength\tabcolsep{2pt}
    \scalebox{1.00}{
    \begin{tabular}{p{3.0cm}  p{1.7cm}<{\centering}p{1.7cm}<{\centering}p{1.7cm}<{\centering}p{1.7cm}<{\centering}p{1.5cm}<{\centering}  p{1.9cm}<{\centering}  p{1.9cm}<{\centering}  p{1.5cm}<{\centering}}
        \toprule
        \multirow{2}{*}{\textbf{Models}} &  \multicolumn{5}{c}{\textbf{Mathematical \& Geometric}} & \textbf{Logical} & \textbf{General} & \multirow{2}{*}{\textbf{Avg.}} \\
        \cmidrule(r){2-6} \cmidrule(r){7-7} \cmidrule(r){8-8}
         &  \textbf{MathVerse} & \textbf{We-Math} & \textbf{MathVision} & \textbf{DynaMath} & \textbf{Geo3k} & \textbf{LogicVista} & \textbf{MMMU-Pro} &  \\
        \midrule
        ThinkLite-VL-7B   & 42.50 & 65.17 & 27.34 & 47.44 & 37.77 & 39.15 & 28.53 & 41.13  \\
        OpenVLThinker-7B  & 41.39 & 65.63 & 25.90 & 51.41 & 38.60 & 43.85 & 32.33 & 42.73  \\
        NoisyRollout-7B  & 39.06   & 63.76   & 24.15   & 53.86   & 42.80   & 46.25   & 35.13   & 43.57   \\
        MM-Eureka-7B & 44.56   & 64.18   & 27.89   & 49.80   & 39.62   & 47.32   & 32.11   & 43.64   \\
        Perception-R1-7B   & 42.99   & 69.05   & 25.42   & 52.84   & 45.19   & 44.32   & 34.36   & 44.88  \\
        VL-Rethinker-7B  & 44.80   & 67.70   & 29.86   & 52.26   & 39.89   & 45.64   & 36.44   & 45.23   \\
        R1-ShareVL-7B  & 44.49   & 69.81   & 27.87   & 53.24   & 42.62   & 46.81   & 34.81   & 45.66  \\
        PAPO-G-7B   & 44.82   & 66.79   & 27.76   & 52.82   & 40.25   & 46.07   & 36.63   & 45.02   \\
        PAPO-D-7B   & 45.32   & 68.30   & 28.36   & 55.83   & 44.11   & 46.70   & 36.34   & 46.42  \\
        Shuffle-R1-7B   & 45.67   & 69.76   & 29.24   & 54.49   & \underline{47.90} & \underline{47.54} & 36.65   & 47.32  \\
        VPPO-7B   & \underline{46.43} & \underline{70.01} & \underline{30.48} & \underline{57.08} & 45.63   & 46.26   & 37.23   & 47.59  \\
        \rowcolor{yellow!10} 
        TPAE-G-Qwen2.5-7B & 45.50   & 69.07   & 30.02   & 56.59   & 46.73   & 47.20   & \textbf{38.27} & \underline{47.63}  \\
        \rowcolor{yellow!10} 
        TPAE-D-Qwen2.5-7B & \textbf{47.15} & \textbf{71.70} & \textbf{30.56} & \textbf{57.39} & \textbf{48.14} & \textbf{48.55} & \underline{37.45} & \textbf{48.71}  \\
        \midrule
        \rowcolor{yellow!10} 
        TPAE-G-Qwen3-8B &  54.61 & 80.44 & 47.55 & 66.88 & 66.31 & 59.93 & 48.61 & 60.62  \\
        \rowcolor{yellow!10} 
        TPAE-D-Qwen3-8B & 55.14 & 81.10 & 48.28 & 66.70 & 68.18 & 61.07 & 48.95 & 61.35  \\
        \bottomrule
    \end{tabular}
    }
\end{table*}

\subsection{Token-level Advantage Estimation}

We integrate the token's trustworthiness score into the existing RLVR frameworks to refine the rollout-level advantage, effectively penalizing tokens that deviate from the established trust region.
Theoretically, failing to reject $H_0$ indicates that the token's vision-entropy state is statistically aligned with the reference distribution at a 99\% confidence level. 
Since this alignment is a fundamental requirement rather than an optimization target, 
we exclusively impose a penalty on tokens that violate this requirement—specifically those for which $H_0$ is rejected due to $\mathcal{M}_t > \tau$. 
The final token-level advantage is then defined as:
\begin{equation}
\hat{A}^{\text{TPAE}}_{ij} = \hat{A}^{\text{GRPO}}_{ij} + \text{sigmoid}(\min(0,\mathcal{T}_{ij}))-0.5
\end{equation}
where $\mathcal{T}_{ij}=\tau -\mathcal{M}_{t}(\mathtt{z}_{ij})$ and $\hat{A}^{\text{GRPO}}_{ij}$ is defined in Eq.~\ref{eq:adv_grpo}.
The sigmoid function $\sigma(\cdot)$ is applied to constraining the output range since $\mathcal{M}_{t} \in [0,+\infty)$. 
By centering the transformation such that the penalty vanishes when $\mathcal{T}_{ij} \geq 0$, we preserve the original advantage for all compliant tokens while providing a dense, bounded signal to suppress vision-entropy misalignment during reinforcement learning.
A detailed, step-by-step implementation of the entire training procedure is provided in the supplementary material. 

\section{Experiment}

We validate the effectiveness of TPAE in advancing multimodal reasoning by integrating it into two foundational RL algorithms across two model backbones. 
Our evaluation comprises a comprehensive comparative study against state-of-the-art multimodal reasoning models across seven benchmarks.
Experimental results demonstrate significant performance gains attributable to TPAE over base RL algorithms, observed in both training efficiency and downstream accuracy. Furthermore, we include a sensitivity analysis on the trust boundary $\tau$ and a qualitative analysis that illustrates the precision of our method in localizing multimodal reasoning failures.

\begin{table*}[t]
    \centering
    \caption{Performance (avg \@8 acc \%) comparison of multimodal reinforcement learning with (\textit{w/}) and without (\textit{w/o}) TPAE.}
    \vspace{-0.1cm}
    \label{table:compare_with_base}
    \setlength\tabcolsep{2pt}
    \scalebox{0.96}{
    \begin{tabular}{p{3.0cm} p{1.7cm}<{\centering}p{1.7cm}<{\centering}p{1.7cm}<{\centering}p{1.7cm}<{\centering}p{1.5cm}<{\centering}  p{1.9cm}<{\centering}  p{1.9cm}<{\centering}  p{1.5cm}<{\centering}}
        \toprule
        \textbf{Models} &  \textbf{MathVerse} & \textbf{We-Math} & \textbf{MathVision} & \textbf{DynaMath} & \textbf{Geo3k} & \textbf{LogicVista} & \textbf{MMMU-Pro} &  \textbf{Avg.} \\
        \midrule
        Qwen2.5-VL-7B & 28.53 & 46.13 & 18.51 & 45.20 & 36.23 & 42.51 & 25.64 & 34.68 \\
        \cmidrule(r){2-9}
        $\quad$+ GRPO \textit{w/o} TPAE & 42.42 & 65.88 & 27.27 & 51.12 & 42.12 & 43.71 & 34.53 & 43.86 \\
        \rowcolor{yellow!10} 
        $\quad$+ GRPO \textit{w/} TPAE & \textbf{45.50}   & \textbf{69.07}   & \textbf{30.02}   & \textbf{56.59}   & \textbf{46.73}   & \textbf{47.20}   & \textbf{38.27} & \textbf{47.63}  \\
        \cmidrule(r){2-9}
        $\quad$+ DAPO  \textit{w/o} TPAE & 44.22 & 66.18 & 27.14 & 53.12 & 43.11 & 45.86 & 35.19 & 44.97 \\
        \rowcolor{yellow!10} 
        $\quad$+ DAPO \textit{w/} TPAE & \textbf{47.15} & \textbf{71.70} & \textbf{30.56} & \textbf{57.39} & \textbf{48.14} & \textbf{48.55} & \textbf{37.45} & \textbf{48.71}  \\
        \midrule
        Qwen3-VL-8B & 40.98 & 65.58 & 27.15 & 60.40 & 53.83 & 51.48 & 33.86 & 47.61 \\
        \cmidrule(r){2-9}
        $\quad$+ GRPO  \textit{w/o} TPAE & 52.79 & 77.32 & 44.92 & 64.48 & 63.39 & 59.10 & 46.24 & 58.32 \\
        \rowcolor{yellow!10} 
        $\quad$+ GRPO \textit{w/} TPAE & \textbf{54.61} & \textbf{80.44} & \textbf{47.55} & \textbf{66.88} & \textbf{66.31} & \textbf{59.93} & \textbf{48.61} & \textbf{60.62} \\
        \cmidrule(r){2-9}
        $\quad$+ DAPO \textit{w/o} TPAE & 53.08 & 78.30 & 45.44 & 65.06 & 64.35 & 58.98 & 46.76 & 58.85  \\
        \rowcolor{yellow!10} 
        $\quad$+ DAPO \textit{w/} TPAE & \textbf{55.14} & \textbf{81.10} & \textbf{48.28} & \textbf{66.70} & \textbf{68.18} & \textbf{61.07} & \textbf{48.95} & \textbf{61.35} \\
        \bottomrule
    \end{tabular}
    }
\end{table*}

\subsection{Experimental Setup}

\subsubsection{Models and Baselines} 
We validate our method using Qwen2.5-VL-7B-Instruct \cite{qwen2_5-VL} and Qwen3-VL-8B-Instruct \cite{qwen3technicalreport} as the base models, integrated within the standard RL algorithms GRPO and DAPO.
To ensure a comprehensive comparison, we benchmark against a diverse suite of leading multimodal reasoning models:
ThinkLite-VL-7B \cite{wang2025sota_Thinklite_VL},
OpenVLThinker-7B \cite{deng2025openvlthinker},
NoisyRollout-7B \cite{liu2025noisyrollout},
MM-Eureka-7B \cite{meng2025mmeurekaexploringfrontiersmultimodal},
Perception-R1-7B \cite{xiao2026perceptionr},
VL-Rethinker-7B \cite{wang2025vlrethinker},
R1-ShareVL-7B \cite{yao2025rsharevl},
PAPO-G-7B, PAPO-D-7B \cite{wang2026perceptionaware},
VPPO-7B \cite{huang2026spotlight},
and Shuffle-R1-7B \cite{zhu2026shuffler1efficientrlframework}.
Notably, all comparative baselines utilize Qwen2.5-VL-7B-Instruct as the base model.

\begin{figure*}[t]
    \centering
    \includegraphics[width=0.98\linewidth]{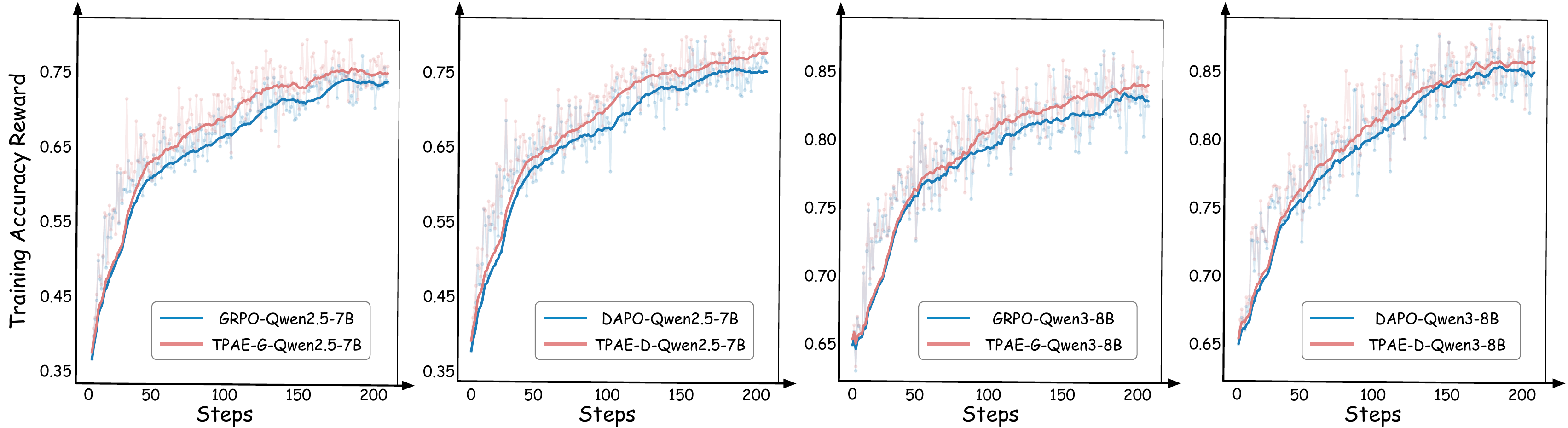}
    \caption{Comparison of RLVR training dynamics on accuracy rewards. Solid lines denote running averages with a window size of 20. TPAE achieves consistently faster learning on both GRPO and DAPO across two model backbones.}
    \label{fig:training_dynamics}
\end{figure*}

\subsubsection{Dataset, Benchmarks, and Evaluation} 
We train our base models on the ViRL39K dataset \cite{wang2025vlrethinker}, which contains 38.9k diverse multimodal reasoning problems.
We conduct a comprehensive evaluation across seven multimodal benchmarks, spanning mathematical, geometric, logical, and multi-discipline reasoning: MathVerse \cite{mathverse_zhangrenrui},  We-Math \cite{wemath_qiaorunqi}, MathVision \cite{mathvision_wangke}, DynaMath \cite{dynamath_zouchengke}, Geo3k \cite{geometry3k_lupan}, LogicVista \cite{logicvista_xiaoyijia}, and MMMU-Pro \cite{yue-etal-2025-mmmu-pro}. 
For MathVerse, we evaluate on the full set of 3.94k multimodal examples, excluding the text-only subset to focus on tasks requiring multimodal context.
For evaluation setup, we employ an exact-match scoring protocol against ground-truth answers, which eliminates the reliance on LLM-as-a-judge systems \cite{wang2026perceptionaware,huang2026spotlight}. 
We report average accuracy@8 at an inference temperature of 1.0, using a single, fixed evaluation pipeline across all models to ensure a rigorous and fair comparison.

\subsubsection{Training Details} 
We develop four model variants using different base RL algorithms and model backbones. Specifically, TPAE-G-Qwen2.5-7B and TPAE-D-Qwen2.5-7B are initialized from Qwen2.5-VL-7B-Instruct and trained using the GRPO and DAPO algorithms, respectively. Following the same paradigm, we extend our method to Qwen3-VL-8B-Instruct to produce TPAE-G-Qwen3-8B and TPAE-D-Qwen3-8B.
The training process utilizes the VerL \cite{hybridflow_verl} reinforcement learning framework for policy optimization. 
Our models are trained for 2 epochs with a learning rate of 1e-6 on 8 NVIDIA Pro6000 96G GPUs.
We employ a training batch size of 384, with 16 rollouts sampled per question.
We set the maximum response length to 2048, and add a small entropy penalty of 0.06 to ensure training stability \cite{wang2026perceptionaware,huang2026spotlight,liu2025noisyrollout}.
Wihin our method, we set the trust region boundary at $\tau=3.03$, corresponding to a 99\% confidence level under the Chi-squared distribution $\chi^2_2$.

We provide more details of benchmark compositions, implementation details and computing resources in the supplementary materials and anonymous GitHub repository.

\begin{table*}[t]
    \centering
    \caption{Ablation study on the trust boundary $\tau$ based on TPAE-G-Qwen2.5-7B.}
    \vspace{-0.1cm}
    \label{table:ablation_tau}
    \setlength\tabcolsep{2pt}
    \scalebox{0.98}{
    \begin{tabular}{p{2.5cm} p{1.7cm}<{\centering}p{1.7cm}<{\centering}p{1.7cm}<{\centering}p{1.7cm}<{\centering}p{1.5cm}<{\centering}  p{1.9cm}<{\centering}  p{1.9cm}<{\centering}  p{1.5cm}<{\centering}}
        \toprule
        \textbf{Configuration} &  \textbf{MathVerse} & \textbf{We-Math} & \textbf{MathVision} & \textbf{DynaMath} & \textbf{Geo3k} & \textbf{LogicVista} & \textbf{MMMU-Pro} &  \textbf{Avg.} \\
        \midrule
        $\tau=2.45$ & 45.15 & 67.99 & 29.30 & 55.72 & 45.74 & 46.76 & \underline{38.09} & 46.96 \\
        $\tau=3.26$ & \underline{45.42} & \underline{68.84} & \underline{29.80} & \underline{55.83} & \textbf{46.80} & \underline{46.84} & 37.87 & \underline{47.34} \\
        \rowcolor{yellow!10} 
        $\tau=3.03$ & \textbf{45.50}   & \textbf{69.07}   & \textbf{30.02}   & \textbf{56.59}   & \underline{46.73}   & \textbf{47.20}   & \textbf{38.27} & \textbf{47.63}  \\
        \bottomrule
    \end{tabular}
    }
\end{table*}

\begin{figure*}[t]
    \centering
    \includegraphics[width=0.98\linewidth]{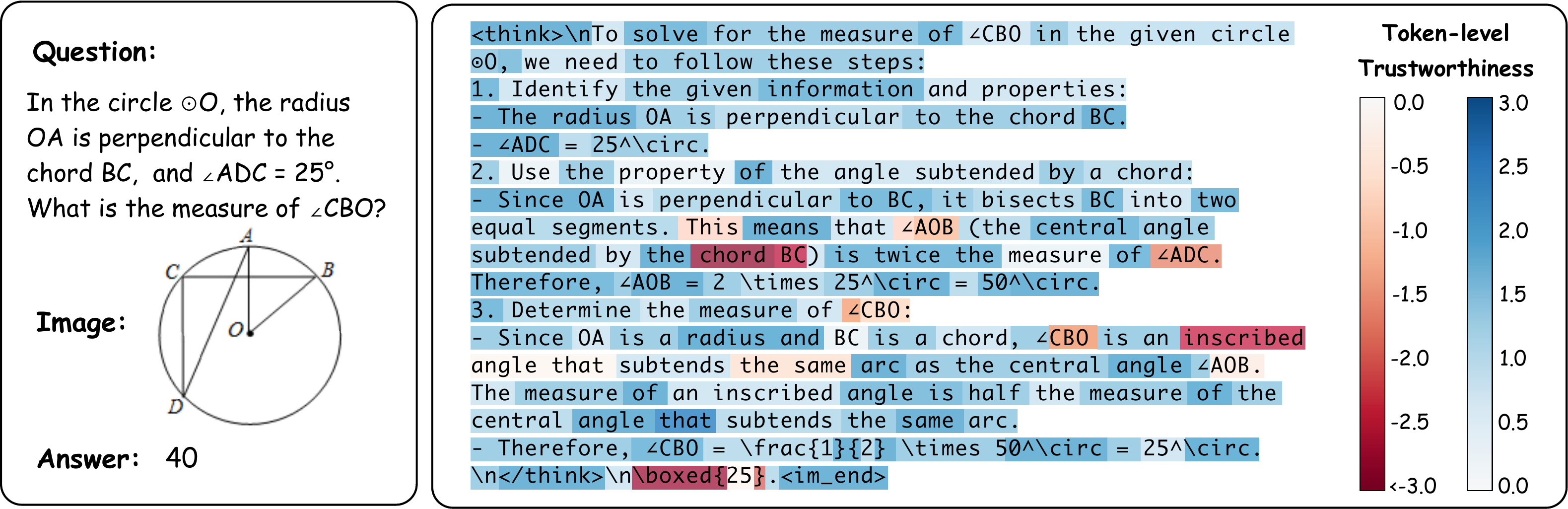}
    \caption{Qualitative analysis of token-level trustworthiness in a failed reasoning trajectory. Dark red highlights successfully localize the pivotal triggers of multimodal reasoning collapse, demonstrating the effectiveness of our TPAE algorithm.}
    \label{fig:qualitative_analysis}
\end{figure*}

\subsection{Main Results}

As illustrated in Table~\ref{table:experimental_result}, TPAE-D-Qwen2.5-7B consistently surpasses current state-of-the-art multimodal reasoning models across all seven multimodal benchmarks. 
By integrating our fine-grained perception-grounded advantage estimation into the DAPO framework, 
TPAE-D-Qwen2.5-7B attains a leading average accuracy of 48.71\% over competitive baselines, outperforming Shuffle-R1-7B by 1.39\% and surpassing VPPO-7B by  1.12\%.  
Similarly, the GRPO-based variant, TPAE-G-Qwen2.5-7B, achieves an average accuracy of 47.63\%, securing the second-highest performance among all models sharing the same Qwen2.5-VL-7B backbone.
The performance advantage of TPAE extends to the Qwen3-VL-8B series, demonstrating its robust scalability across evolving model architectures. Both TPAE-G-Qwen3-8B and TPAE-D-Qwen3-8B variants establish new performance frontiers by achieving an average accuracy of 60.62\% and 61.35\% respectively. The superiority of TPAE is further evidenced by its improved training dynamics relative to base RL algorithms as discussed in Section~\ref{sec:compare_with_base_RL}.

\subsection{Ablation Studies}

\subsubsection{Comparison with Base RL Algorithm}
\label{sec:compare_with_base_RL}
The effectiveness of TPAE is evidenced by its superior training dynamics and  downstream performance compared with standard GRPO and DAPO implementations.
In Figure~\ref{fig:training_dynamics}, we display the training dynamics based on accuracy rewards on ViRL39K. The integration of TPAE yields accelerated initial convergence; specifically, models equipped with TPAE consistently surpass the peak accuracy of base RL algorithms while maintaining identical training configurations, including the training dataset, rollout space, and reward functions.

The integration of TPAE also yields consistent improvements in downstream performance across all benchmarks and model backbones. As shown in Table~\ref{table:compare_with_base}, incorporating TPAE within the DAPO framework leads to average accuracy gains of 3.74\% on Qwen2.5-VL-7B and 2.50\% on Qwen3-VL-8B compared to their respective base RL baselines.
Similarly, TPAE-integrated GRPO optimization achieves improvements of 3.77\% on the 7B variant and 2.30\% on the 8B variant. 
Together, these results suggest that our perception-grounded advantage estimation facilitates a more efficient convergence toward superior reasoning capabilities.

\subsubsection{Sensitivity of Trust Boundary $\tau$}
\label{sec:tau_ablation}
We perform a sensitivity analysis on the hyperparameter $\tau$ in Eq.~\ref{eq:trust}, which determines the trust region for detecting token-level vision-entropy misalignment relative to the reference distributions.
These experiments are conducted on Qwen2.5-VL-7B-Instruct utilizing GRPO as the base RL algorithm.
We compare our default $\tau=3.03$, representing a 99\% confidence level, against a more permissive $\tau=2.45$ at 95\% confidence and a more stringent $\tau=3.26$ at 99.5\% confidence.
As shown in Table~\ref{table:ablation_tau}, the default setting $\tau=3.03$ yields the highest accuracy on six out of seven benchmarks.
While increasing the threshold to $\tau=3.26$ results in a slight decrease in average accuracy to 47.34\%, a more permissive threshold of $\tau=2.45$ leads to a more notable degradation to 46.96\%. 
These results suggest that a 95\% confidence level is insufficiently rigorous for isolating tokens responsible for reasoning failures. Conversely, our choice of $\tau=3.03$ provides a more effective filtering mechanism that precisely targets tokens responsible for multimodal reasoning failures.

\subsection{Qualitative Analysis}

We provide a qualitative analysis in Figure \ref{fig:qualitative_analysis} to illustrate the token-level trustworthiness defined in Eq. \ref{eq:trust}. Tokens with high untrustworthiness are highlighted in dark red and are concentrated at visually-hallucinated reasoning steps.
The reasoning collapse begins when the model erroneously identifies $\angle AOB$ as the central angle subtending the arc $BC$. By failing to recognize that $\angle AOB$ actually subtends arc $AB$, the model establishes an incorrect geometric association with $\angle ADC$.
This is compounded when the model incorrectly treats $\angle CBO$ as an inscribed angle rather than an interior angle of the isosceles triangle $\triangle OBC$.
Notably, the final output syntax, including the \texttt{boxed} terminal, also exhibits elevated untrustworthiness \cite{sun2025ktae}.

\section{Conclusion}

In this paper, we pioneer an exploration of multimodal reinforcement learning through the lens of token-level vision-entropy dynamics.
Our preliminary study reveals that the joint distribution of visual dependency and predictive entropy establishes an effective indicator for multimodal reasoning collapse. Leveraging this insight, we propose \textbf{T}oken-level \textbf{P}erception-grounded  \textbf{A}dvantage \textbf{E}stimation (\textbf{TPAE}), a novel algorithm that derives fine-grained advantages by statistically aligning a token's vision-entropy state with the reference distribution observed in correct reasoning chains. TPAE not only establishes a new state-of-the-art across a diverse suite of multimodal benchmarks but also substantially accelerates convergence throughout the RL process. We believe TPAE represents a pioneering step toward perception-aware policy optimization, establishing a theoretically grounded approach for scaling the reasoning capabilities of MLLMs within the RLVR frameworks.

\section*{Acknowledgments}
This research is supported by the National Research Foundation, Singapore under its National Large Language Models Funding Initiative (AISG Award No: AISG-NMLP-2024-002). This research is also supported by the Ministry of Education, Singapore, under its AcRF Tier 2 Funding (Proposal ID: T2EP20123-0052). Any opinions, findings, conclusions, or recommendations expressed in this material are those of the author(s) and do not reflect the views of the National Research Foundation or the Ministry of Education, Singapore.

\bibliographystyle{ACM-Reference-Format}
\balance
\bibliography{main}

\end{document}